\documentclass{svproc}
\usepackage{url}

\usepackage{tikz}
\usepackage{amsmath}

\usepackage{filecontents}

\usepackage{multirow}
\usepackage{algorithmic}
\usepackage{algorithm}
\usepackage{enumitem}
\usepackage{graphicx}
\usepackage{booktabs}
\usepackage{eqparbox}
\usepackage{arydshln}
\usepackage{arydshln}
\usepackage{multirow}
\usepackage{amsmath}
\usepackage{array}
\usepackage{float}
\usepackage{soul}
\usepackage{microtype}
\usepackage{latexsym}
\usepackage{adjustbox}

\usepackage[final]{listofsymbols}
\opensymdef
\newsym[True anomaly $({}^\circ)$]{trueanomaly}{\Theta}
\closesymdef

\begin{document}
\mainmatter              
\title{Prompt-Based Learning for \\Thread Structure Prediction \\in Cybersecurity Forums}
\titlerunning{Prompt-Based Thread Structure Prediction}  
%
\author{Kazuaki Kashihara\inst{1} \and Kuntal Kumar Pal\inst{1} \and  Chitta Baral\inst{1} \and Robert P Trevino\inst{2} }
%
\authorrunning{Kazuaki Kashihara et al.}
%
\tocauthor{Kazuaki Kashihara, Kuntal Kumar Pal, Chitta Baral, Robert P Trevino}
%
\institute{School of Computing and Augmented Intelligence, Arizona State University, USA\\
\email{(kkashiha,kkpal,chitta)@asu.edu}
\and
Design Pickle, USA\\
\email{rtrevino@designpickle.com}
}

\maketitle              

\begin{abstract}
With recent trends indicating cyber crimes increasing in both frequency and cost, it is imperative to develop new methods that leverage data-rich hacker forums to assist in combating ever evolving cyber threats. Defining interactions within these forums is critical as it facilitates identifying highly skilled users, which can improve prediction of novel threats and future cyber attacks. We propose a method called Next Paragraph Prediction with Instructional Prompting (NPP-IP) to predict thread structures while grounded on the context around posts.
This is the first time to apply an instructional prompting approach to the cybersecurity domain. We evaluate our NPP-IP with the Reddit dataset and Hacker Forums dataset that has posts and thread structures of real hacker forums' threads, and compare our method's performance with existing methods.
The experimental evaluation shows that our proposed method can predict the thread structure significantly better than existing methods allowing for better social network prediction based on forum interactions. 

\keywords{Instructional prompts, Thread structure prediction, Thread Structure, Social network, Unstructured forums, Cybersecurity}
\end{abstract}

\section{Introduction}
\begin{figure}[t!]
    \centering
    \includegraphics[width=.95\textwidth]{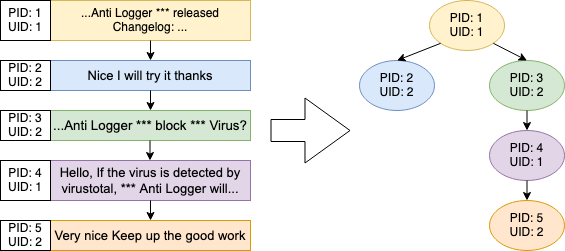}
    \caption{Example of a thread from an unstructured Hacker forum in the darkweb (left), and the predicted thread structure (right).}
    \label{fig:structur}
\end{figure}

Cybercrimes cost trillions of dollars in damages worldwide each year, impacting different sectors of society ranging from national defense to private industry~\cite{CSS2021}.  Current trends indicate a considerable rise in cybercrimes in the next several years as hacker tools become more sophisticated and ubiquitous~\cite{cc-report2016}. This is, in part, due to the advent of the dark web, which has given hackers the opportunity to interact, profit, and exchange information on dark web forums~\cite{goel2011cyberwarfare}. Identifying key user interactions within these dark forums can assist in identifying prominent hackers with knowledge of novel threats as well as predicting potential cyber attacks. Thus, the thread structure of a forum becomes important in generating social networks based on user interactions as shown in Fig.~\ref{fig:structur}~\cite{isi/FuAC07}. 

Unfortunately, most of the hacker forums are unstructured making it difficult to identify user interactions through post replies in an automated manner. Moreover, although many of these dark forums have a rich source of text information in different threads that discuss specific topics, such as malware, virus, illegal items, and other illegal activities, the recent reports indicate that $90\%$ of posts on popular dark web hacking forums are made by those looking to solicit hacker services instead of the hackers themselves~\cite{ptsecurity-CHS2021,techtarget-article2021}. Traditional methods used to define social networks on unstructured forums such as Creator-oriented Network and Last Reply-oriented Network~\cite{sigkdd/LHuillierARA10} are based upon temporal interaction assumptions that do not consider the full context of the user interactions based on the content of the posts. Kashihara et al. previously introduced a powerful deep learning method called Next Paragraph prediction (NPP) designed to define social networks using posts from the Reddit forums~\cite{kashihara2020social}. The NPP method outperformed traditional methods as well as BERT's Next Sentence Prediction (NSP) \cite{BERT-DevlinCLT19} when defining social networks from posts.

Building upon the NPP method, we propose the Next Paragraph Prediction with Instructional Prompting (NPP-IP) that leverages cutting-edge Prompt-based learning to assist in social network construction from posts. Using the Reddit dataset~\cite{kashihara2020social} consisting of over $105$ threads and $1,648$ posts, we train and evaluate the model against both traditional methods as well as the original NPP method. In addition, we test the model using real unstructured hacker forums data, where $20$ threads are manually annotated by human experts to identify interactions based on posts. 
Since the hacker forums data is hard to extract and manual annotation takes time and labor tense, we create the above limited dataset for the hacker forums in this work.
The results show that NPP-IP performs $2.68$ to $4.70$ percentage points better than the other existing methods.

\noindent
\textbf{Contributions}: Our research contributions in this research are listed as follows:
\begin{itemize}
    \item Prompt-based learning (instructional prompting) is introduced into a deep learning method called Next Paragraph Prediction for social network construction of forum data.
    \item We apply Prompt-based learning to Cybersecurity domain for the first time.
    \item  The evaluation results show that our method can predict thread structures better than the existing methods.
    \item The results indicate that the proposed method is robust enough to train using data from one cyber-related forum and apply to another cyber-related forum.
\end{itemize}
\section{Related Work}

In this section, we give a brief overview of previous work on reply relationships identification, thread structure prediction, and instructional prompting.

\subsection{Reply Relationships Identification}
In general, reply relationships are explicitly defined in several social platforms such as Reddit, Twitter, and Facebook. However, unstructured forums and some of the online communication tools such as Telegram and other instant messaging platforms have a challenge to identify the reply relationships since users usually do not use explicit reply marks but directly post related messages or posts to communicate. Thus, several methods are proposed to identify the reply relationships in various platforms.

The first step of the conversation disentanglement task is usually reply relationship identification~\cite{acl/ElsnerC08,ijcnlp/MehriC17,naacl/JiangCCW18}. 
There are several classification models that have been developed to detect reply relationships between pairs of messages. A linear binary classifier considering conversation features and content features is proposed by Elsner et al.~\cite{acl/ElsnerC08}.
A random forest classifier with 250 trees that uses a feature vector in order to describe the relationship between the two messages is developed by Mehri et al.~\cite{ijcnlp/MehriC17}.

Neural models are used in recent methods to represent utterances. The joint model of reply relationship and pairwise relationship under pointer network model is proposed by Yu et al.~\cite{emnlp/YuJ20}.
Another approach is that the problem of conversation structure modeling is defined as identifying the parent utterances to which each utterance in the conversation responds to by Zhu et al.~\cite{aaai/ZhuNWNX20}. They designed a novel masking mechanism using masked hierarchical transformer.
Zhang et al.~\cite{zhang2021identifying} proposed a method based on multi-features to identify reply relationships from Telegram groups using BERT model to learn the textual representation of messages and utilize the user's contextual features that contain richer information to overcome the limitation of short messages. 


\subsection{Thread Structure Prediction} 
\begin{figure}[t]
    \centering
    \includegraphics[width=0.95\textwidth]{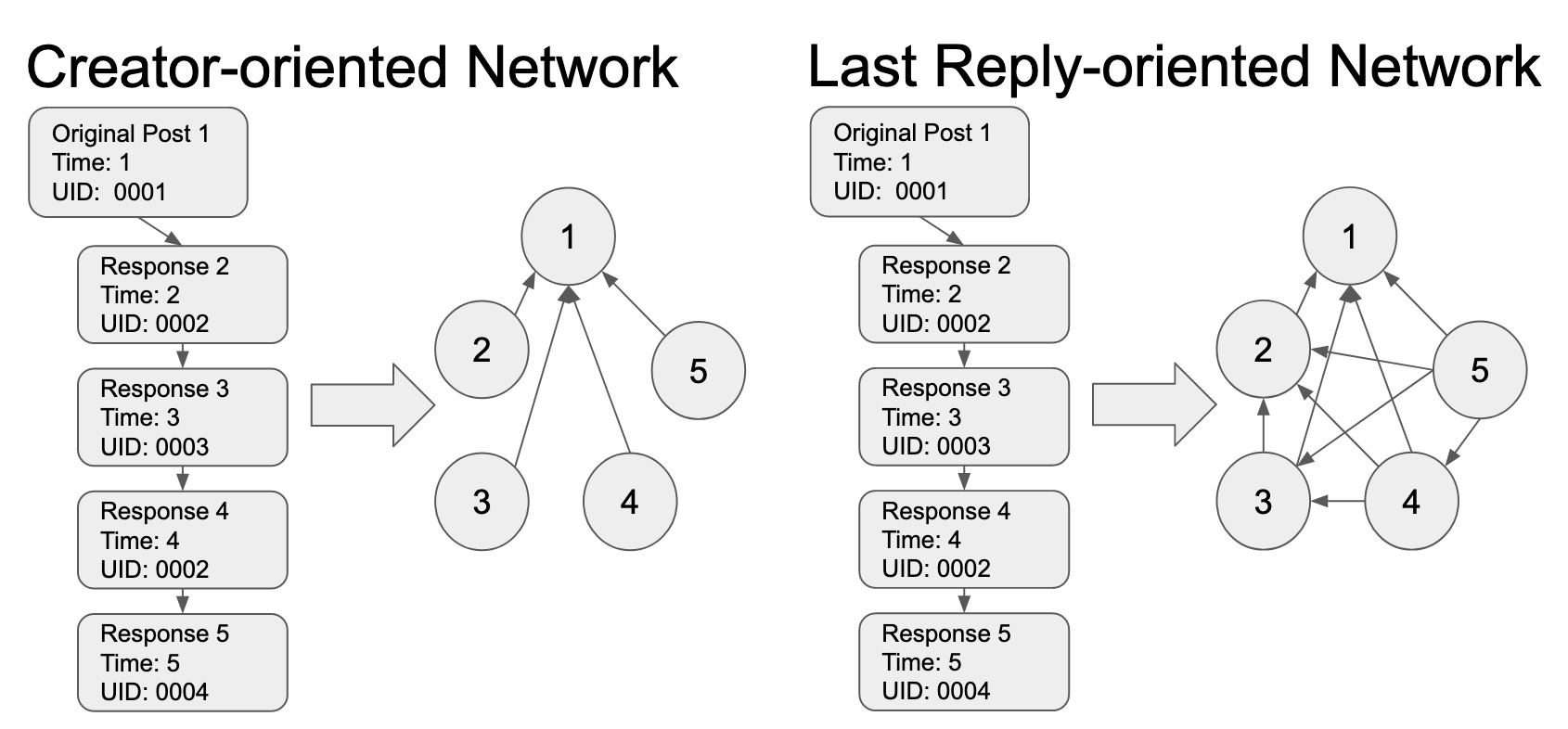}
    \caption{Example of Creator-oriented Network and Last Reply-oriented Network.}
    \label{fig:networkSample}
\end{figure}

In order to build social networks from forums, member interactions must be correctly identified via posts on threads. 
There are two network representations introduced~\cite{sigkdd/LHuillierARA10} for building the social network in forums: Creator-oriented Network and Last Reply-oriented Network.
The Last Reply-oriented Network is widely used for the social network analysis in the recent works~\cite{phillips2015extracting,almukaynizi2017predicting,MarinSS18,SarkarASS18,eurosp/PeteHCB20,isi/JohnsenF20}.
Fig.~\ref{fig:networkSample} shows the sample structure of Creator-oriented Network and Last Reply-oriented Network.
Since these two traditional network conversion approaches are based on limited information and considerable assumptions on interactions between users, the social structures of the networks are likely not accurate representations. Other recent works have predicted helpful posts in the forums~\cite{HalderKS19} using a neural network based model that determines whether the post is useful or not. However, the importance of a post has very little utility when predicting interactions and thus social networks. 
More recently, Kashihara et al. ~\cite{kashihara2020social} proposed the Next Paragraph Prediction (NPP) method which extended BERT's Next Sentence Prediction to predict the response post from the previous post. This method allows for the Reconstruction of social networks using thread structure prediction.

\subsection{Instructional Prompting}
Building effective discrete prompts for language models (LM) to perform NLP tasks is an active area of research~\cite{emnlp/SchickS21,naacl/ScaoR21,emnlp/TamMBSR21,chi/ReynoldsM21}. Such prompts are often extremely short and may not include a complete definition of complex tasks. In contrast, the recent works~\cite{corr/abs-2109-07830,corr/abs-2104-08773} give instructions encode detailed instructions as they were used to collect the dataset. Driven many empirical analysis by~\cite{corr/abs-2109-07830}, the framing instructional prompting has demonstrated considerable improvements across LMs. 


\section{Model Description}

\begin{figure}[t!]
    \centering
    \includegraphics[width=.95\textwidth]{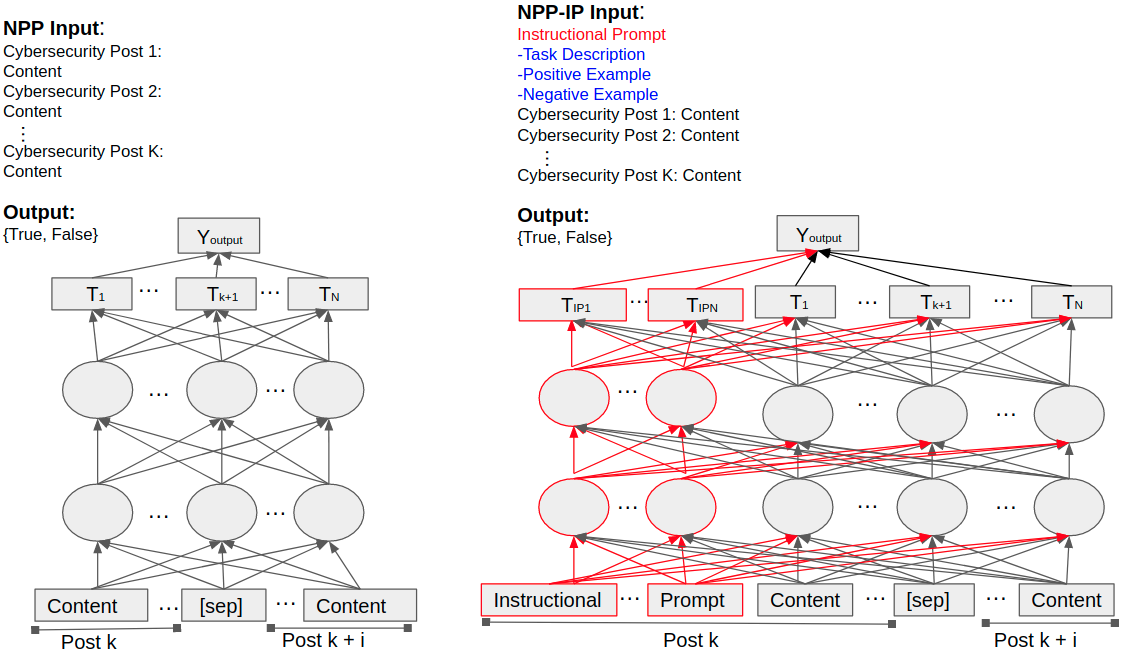}
    \caption{The original NPP model (Left) combines a pair of posts to predict whether one post is a response to the other. Our NPP-IP model (Right) incorporates instruction prompt information into the NPP structure allowing for task information to be leveraged.}
    \label{fig:dataset}
\end{figure}

 Our proposed NPP-IP model is based on infusing the original dataset with specific task instructions using an instruction prompting function. Formally, the instruction prompting function $f_{prompt}(\cdot)$ is defined as 
\begin{align}
    f_{prompt}(x)  =  I || x, 
\end{align}
where || represents concatenation of instruction prompt $I$ with training sample $x$.  Instruction prompt $I$ is formally defined as:\\

	\parbox{4.25in}{
    \small \textbf{Task Description:}\\
	You are given two posts and you need to generate True if they are the direct reply relation, otherwise generate False.\\
	\textbf{Positive Example:}\\
    post1: Windows Defender Gets a New Name: Microsoft Defender\\
    post2: Bring back MSE and its ui even logo looks cool...\\
    output: True\\
    \textbf{Negative Example:}\\
    post1: Windows Defender Gets a New Name: Microsoft Defender\\
    post2: Title says it\\
    output: False"
    \\}
    
Training sample $x$ is formally defined as
\begin{align}
    x = \text{Post $k$ || [sep] || Post $k+i$},
\end{align}
which represents a pair of concatenated posts at index $k$ and $k+i$ with a separation key $[sep]$, such that  $i\neq0$.

The NPP-IP model leverages five framing techniques defined in \cite{corr/abs-2109-07830} for framing the instruction prompting information $I$. (i) First, the \textbf{Use Low Level Patterns} technique is accomplished by providing a simple task descriptor to correctly output a value of True or False if a reply relationship exists between posts without including any cybersecurity jargon. (ii and iii) Second, \textbf{Itemized Instructions} are provided via the positive and negative examples with the corresponding output in bulleted list format for thread structure prediction. The positive and negative examples also fulfill the \textbf{Break It Down} technique by defining simpler sub-tasks corresponding to identifying negative and positive examples. This is also where cybersecurity information is introduced into the instructional prompt. (iv) Next, \textbf{Enforce Constraints} is accomplished by constraining the examples to their respective outputs of True or False. (v) Lastly, the \textbf{Specialize Instructions} technique is accomplished by specifically stating the expected output in both task description and examples.

\begin{figure}[t!]
    \centering
    \includegraphics[width=.6\textwidth]{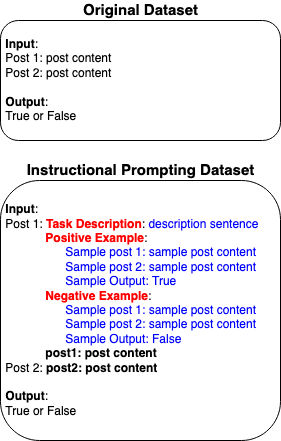}
    \caption{The data structure for NPP model and NPP-IP model. }
    \label{fig:datasetStructure}
\end{figure}

Fig.~\ref{fig:dataset} shows the BERT-based neural network structure used by an NPP model as well as the resultant NPP-IP model after introducing instructional prompting information. The original dataset gives two posts as its input, where the label space is defined as \{True, False\}, defining whether posts share a direct response relation or not. Including instructional prompting provides critical task information for both positive and negative cases, which are then used in the embedding and subsequent prediction task during training. 
Fig.~\ref{fig:datasetStructure} shows the data structure of the model for original NPP model and NPP-IP model.

\section{Evaluation and Results}



\subsection{Datasets} 
A curated Reddit dataset \cite{kashihara2020social} was used to train and evaluate our proposed model. The Reddit dataset is ideally suited for thread structure analysis given its tree-like structure within different threads.
The Reddit dataset has the threads from the following ten topics from ``cybersecurity'' field and extracted the threads of these topics: ``cyber\_security'', ``AskNetsec'', ``ComputerSecurity'', ``cyberpunk'', ``cybersecurity'', ``Hacking'', ``Malware'', ``Malwarebytes'', and ``security''. 
Our proposed model was also evaluated using 20 hacker forum threads from three English hacker forums annotated by human experts, which is referred to as the ``Hacker Forums" dataset. The forum thread data is from CYR3CON (Cyber Security Works)\footnote{\url{https://www.cyr3con.ai}}. The average posts per thread is $15.4$. Four cybersecurity experts checked posts in each thread, and annotated a relation of two posts in a thread which the two posts are direct response relations or not. The site names and usernames are anonymized. 
Table~\ref{tab:stat} and Table~\ref{tab:statHF} show the basic statistics of Reddit and Hacker Forums dataset respectively.
In the Reddit dataset, we create all pairs of post combinations in a thread. The positive pair of posts is that the second post is the direct reply of the first post. There are 14744 pairs of posts including both positive and negative. Then, we split the pairs into train, dev, and test sets and the split ratio is $60\%$, $10\%$, and $30\%$ respectively.

\begin{table}[t!]
    \centering
    \begin{tabular}{|c|c|c|}
        \hline Topic Name & TH & Posts \\
        \hline
        \hline cyber\_security & 8 & 48  \\
        \hline AskNetsec & 14 & 338 \\
        \hline ComputerSecurity & 12 & 110 \\
        \hline cyberpunk & 11 & 176  \\
        \hline cybersecurity & 11 & 158  \\
        \hline Hacking & 12 & 370  \\
        \hline Hacking\_Tutorial & 12 & 110 \\
        \hline Malware & 9 & 82 \\
        \hline Malwarebytes & 8 & 72  \\
        \hline security & 8 & 184  \\
        \hline
    \end{tabular}
    \caption{The Reddit dataset consisted of ten cybersecurity topics. ``TH" is defined as the number of threads in each topic while ``Posts" is defined as the number of Posts across the different threads.}
    \label{tab:stat}
\end{table}

\begin{table}[t!]
    \centering
    \begin{tabular}{|c|c|c|}
        \hline Forum \# & TH & Posts \\
        \hline
        \hline Forum1 & 7 & 169  \\
        \hline Forum2 & 7 & 80 \\
        \hline Forum3 & 6 & 58 \\
        \hline
    \end{tabular}
    \caption{The Hacker Forums dataset consisted of 20 threads from three hacker forums. ``TH" is defined as the number of threads in each topic while ``Posts" is defined as the number of Posts across the different threads.}
    \label{tab:statHF}
\end{table}


\begin{table}[t!]
\centering
\begin{tabular}{|l|l|l|l|l|}
\hline
Method                  & Model & P    & R    & F1            \\ \hline
CO                      & -     & 0.00 & 1.00 & 0.01          \\ \hline
LR                      & -     & 0.72 & 0.12 & 0.20          \\ \hline
\multirow{4}{*}{NPP}    & BE-B  & 0.42 & 0.46 & 0.44          \\ \cline{2-5} 
                        & BE-L  & 0.36 & 0.51 & 0.42          \\ \cline{2-5} 
                        & RB-B  & 0.59 & 0.33 & 0.43          \\ \cline{2-5} 
                        & RB-L  & 0.41 & 0.58 & 0.48
                        \\ \hline
\multirow{4}{*}{NPP-IP} & BE-B  & 0.48 & 0.46 & 0.47 \\ \cline{2-5} 
                        & BE-L  & 0.64 & 0.41 & 0.50 \\ \cline{2-5} 
                        & RB-B  & 0.62 & 0.43 & \textbf{0.51} \\ \cline{2-5} 
                        & RB-L  & 0.39 & 0.56 & 0.46
                        \\ \hline
\end{tabular}
\caption{Results from the Reddit test data show that the NPP-IP method outperformed all other methods for thread structure prediction across all but one of the different BERT language models analyzed.}
\label{table:redditResult}
\end{table}

\begin{table}[t!]
\centering
\begin{tabular}{|l|l|lll|lll|lll|}
\hline
                        &       & \multicolumn{3}{c|}{Forum1}                                           & \multicolumn{3}{c|}{Forum2}                                           & \multicolumn{3}{c|}{Forum3}                                           \\ \hline
Method                  & Model & \multicolumn{1}{l|}{P}    & \multicolumn{1}{l|}{R}    & F1            & \multicolumn{1}{l|}{P}    & \multicolumn{1}{l|}{R}    & F1            & \multicolumn{1}{l|}{P}    & \multicolumn{1}{l|}{R}    & F1            \\ \hline
CO                      & -     & \multicolumn{1}{l|}{0.31} & \multicolumn{1}{l|}{1.00} & 0.47          & \multicolumn{1}{l|}{0.27} & \multicolumn{1}{l|}{1.00} & 0.43          & \multicolumn{1}{l|}{0.12} & \multicolumn{1}{l|}{1.00} & 0.21          \\ \hline
LR                      & -     & \multicolumn{1}{l|}{0.50} & \multicolumn{1}{l|}{0.00} & 0.01          & \multicolumn{1}{l|}{0.50} & \multicolumn{1}{l|}{0.09} & 0.16          & \multicolumn{1}{l|}{0.50} & \multicolumn{1}{l|}{0.13} & 0.21          \\ \hline
\multirow{4}{*}{NPP}    & BE-B  & \multicolumn{1}{l|}{0.40} & \multicolumn{1}{l|}{0.37} & 0.39          & \multicolumn{1}{l|}{0.37} & \multicolumn{1}{l|}{0.61} & 0.46          & \multicolumn{1}{l|}{0.33} & \multicolumn{1}{l|}{0.65} & 0.44          \\ \cline{2-11} 
                        & BE-L  & \multicolumn{1}{l|}{0.94} & \multicolumn{1}{l|}{0.27} & 0.41          & \multicolumn{1}{l|}{0.50} & \multicolumn{1}{l|}{0.34} & 0.41          & \multicolumn{1}{l|}{0.50} & \multicolumn{1}{l|}{0.33} & 0.40          \\ \cline{2-11} 
                        & RB-B  & \multicolumn{1}{l|}{0.59} & \multicolumn{1}{l|}{0.38} & 0.46          & \multicolumn{1}{l|}{0.29} & \multicolumn{1}{l|}{0.50} & 0.37          & \multicolumn{1}{l|}{0.48} & \multicolumn{1}{l|}{0.43} & 0.41          \\ \cline{2-11} 
                        & RB-L  & \multicolumn{1}{l|}{0.54} & \multicolumn{1}{l|}{0.55} & \textbf{0.54}          & \multicolumn{1}{l|}{0.55} & \multicolumn{1}{l|}{0.58} & 0.54          & \multicolumn{1}{l|}{0.45} & \multicolumn{1}{l|}{0.40} & 0.41          \\ \hline
\multirow{4}{*}{NPP-IP} & BE-B  & \multicolumn{1}{l|}{0.55} & \multicolumn{1}{l|}{0.39} & 0.45          & \multicolumn{1}{l|}{0.71} & \multicolumn{1}{l|}{0.63} & \textbf{0.67} & \multicolumn{1}{l|}{0.61} & \multicolumn{1}{l|}{0.56} & \textbf{0.58} \\ \cline{2-11} 
                        & BE-L  & \multicolumn{1}{l|}{0.70} & \multicolumn{1}{l|}{0.43} & 0.53 & \multicolumn{1}{l|}{0.85} & \multicolumn{1}{l|}{0.31} & 0.45          & \multicolumn{1}{l|}{0.61} & \multicolumn{1}{l|}{0.60} & 0.57          \\ \cline{2-11} 
                        & RB-B  & \multicolumn{1}{l|}{0.50} & \multicolumn{1}{l|}{0.37} & 0.42          & \multicolumn{1}{l|}{0.52} & \multicolumn{1}{l|}{0.58} & 0.48          & \multicolumn{1}{l|}{0.53} & \multicolumn{1}{l|}{0.84} & 0.46          \\ \cline{2-11} 
                        & RB-L  & \multicolumn{1}{l|}{0.50} & \multicolumn{1}{l|}{0.87} & 0.43          & \multicolumn{1}{l|}{0.50} & \multicolumn{1}{l|}{0.34} & 0.41          & \multicolumn{1}{l|}{0.50} & \multicolumn{1}{l|}{0.33} & 0.40          \\ \hline
\end{tabular}
\caption{Results from each of the anonymous hacker forums demonstrated that the NPP-IP model outperformed all other models. The NPP and NPP-IP models were both trained with Reddit data further demonstrating NPP-IPs inference performance robustness on unrelated cyber forums.}
\label{table:hfFullResults}
\end{table}

\subsection{Metrics and Task} 
\begin{figure}[t!]
    \centering
    \includegraphics[width=.8\textwidth]{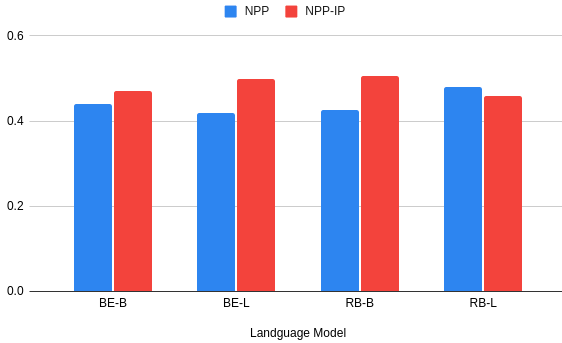}
    \caption{An F1 score comparison of the Reddit Forum data using different BERT-based language models indicates that our proposed NPP-IP model (Red) outperforms NPP (Blue) in all but one language model, achieving the highest score using RoBERTa LM.}
    \label{fig:LMcomp}
\end{figure}

Our proposed NPP-IP method was evaluated against several different methods for thread structure prediction using cybersecurity related posts. Two language models, BERT (BE) and RoBERTa (RB), were explored when training the NPP and proposed NPP-IP models, where -B and -L represent base and large models for each LM respectively. As shown in Fig. \ref{fig:LMcomp}, our NPP-IP method outperformed the original NPP method based on the F1 score using the Reddit data across all but one of the LMs.  
 We compared performance with well known methods, Creator-Oriented Network (CO) and Last Reply-Oriented Network (LR), using Precision (P), Recall (R), and F1 score (F1) metrics.

\subsection{Libraries and Hyperparameters} In order to build, train, and evaluate both NPP and NPP-IP methods, we use publicly available libraries and set hyperparameters, and they are described as follows: torchtext 0.8.0 and PyTorch 1.7.1~\cite{paszke2017automatic}, pytorch-lightning 1.2.2~\cite{falcon2020framework}, and transformers 3.4~\cite{emnlp/WolfDSCDMCRLFDS20} on Google Colab (Nvidia K80 12 GB GPU) were used. We use the hidden dropout probability as 0.15. The batch size was set to 8 and the learning rate was set to 5e-6. The model was trained with $>10$ epochs. Convergence was observed around 3 epochs with limited over-fitting and the maximum sequence length was set to 250.



\subsection{Results}
 The results are reported in Tables~\ref{table:redditResult} and \ref{table:hfFullResults} for Reddit and Hacker Forums datasets, respectively. Both tables show a clear improvement from our NPP-IP method compared to most methods across both datasets. There were two cases where NPP outperformed NPP-IP with both using the large RoBERTa (RB-L) language model. 
As shown in Table~\ref{table:hfFullResults}, in two of the three hacker forums, our proposed NPP-IP method with BERT-B LM reached the highest F1 score while NPP-IP with BERT-L LM recorded the highest F1 score for the third forum. 


\section{Analysis and Discussion}
In this section, we analysis the results of Reddit and Hacker Forums datasets, especially NPP-IP performance, Precision and Recall Implications, and various Error cases.

\subsection{NPP-IP Performance}
As far as the authors are aware, this is the first time that instructional prompts have been applied to text-based cybersecurity data. As the results of Reddit and Hacker Forums datasets show in Table~\ref{table:redditResult} and Table~\ref{table:hfFullResults}, NPP-IP performs better than NPP in most of the cases. The improvement on the Reddit dataset from NPP to NPP-IP across different LMs, ranged between $3-8\%$ across F1 scores. Similar improvements were observed using the real world Hackers Forums dataset, ranging between $3-11\%$ difference across F1 scores. This is despite the fact that the models were trained on Reddit only data. As shown, in Table \ref{tab:stat}, the Reddit dataset is comprised of cybersecurity related topics across the different threads. \textbf{This evidence is consistent with NPP-IP's ability to better detect and leverage cybersecurity related information compared to other well known methods for social network construction based on thread structure prediction.} Moreover, the framing of the instructional prompt using cyber related information may also have improved its performance across different forums. This is a significant discovery since annotating new datasets, especially in the cyber realm, is costly, requiring considerable human experts' efforts to collect a decent size of data for training and testing. More research needs to be conducted to determine the extent to which framing cyber related instruction prompts can make text-based analysis more robust across different cyber forums and datasets.


\subsection{Precision and Recall Implications}
As precision and recall scores in Tables~\ref{table:redditResult} and ~\ref{table:hfFullResults} show, both are considerably low in Reddit dataset, with recall scores much lower than precision scores in the Hacker Forums dataset. One possible explanation for this observed behavior is that publicly available pre-trained LMs were used. These LMs are pre-trained by a wide range of topics across a massive size of data. However, the cybersecurity field is in a constant state of flux - changing the meaning of words and adding new words quite frequently. We suspect that LMs could not understand many of the cybersecurity keywords in posts predicting thread structures that were consistent with actual social interaction. Thus, re-training LMs with cybersecurity data should be explored to improve the performance.

\subsection{Error Cases}
\begin{figure}[t!]
    \centering
    \includegraphics[width=.5\textwidth]{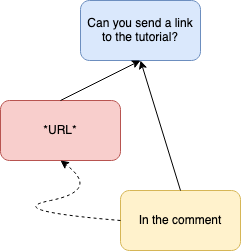}
    \caption{An interesting case in Reddit dataset. Since the actual URL is harmful site URL, we replaced it as *URL*. }
    \label{fig:reddit}
\end{figure}

There are several error cases that are not easy to solve.
Fig.~\ref{fig:reddit} shows an interesting case we found in the Reddit dataset. For the question post ``Can you send a link to the tutorial?'', a user responded ``*URL*'' and ``in the comment''. In the ground truth, ``in the comment'' is the response of ``*URL*'', however, both of our models predicted ``in the comment'' is the response of ``Can you send a link to the tutorial?''. We think that ``in the comment'' reinforces the post ``*URL*'' and also answers the original question. Since the ground truth is based on the thread tree structure, it only has one interaction even if it can interact with multiple posts or users. However, due to the tree structure in Reddit, the ground truth from the subreddit structure is assigned to only one of them. We found some cases that our methods predicted a post replied to multiple posts, and only one of them is correct as we mentioned before. Thus, these cases may decrease the performance of our methods.

\begin{figure}[t]
    \centering
    \includegraphics[width=.8\textwidth]{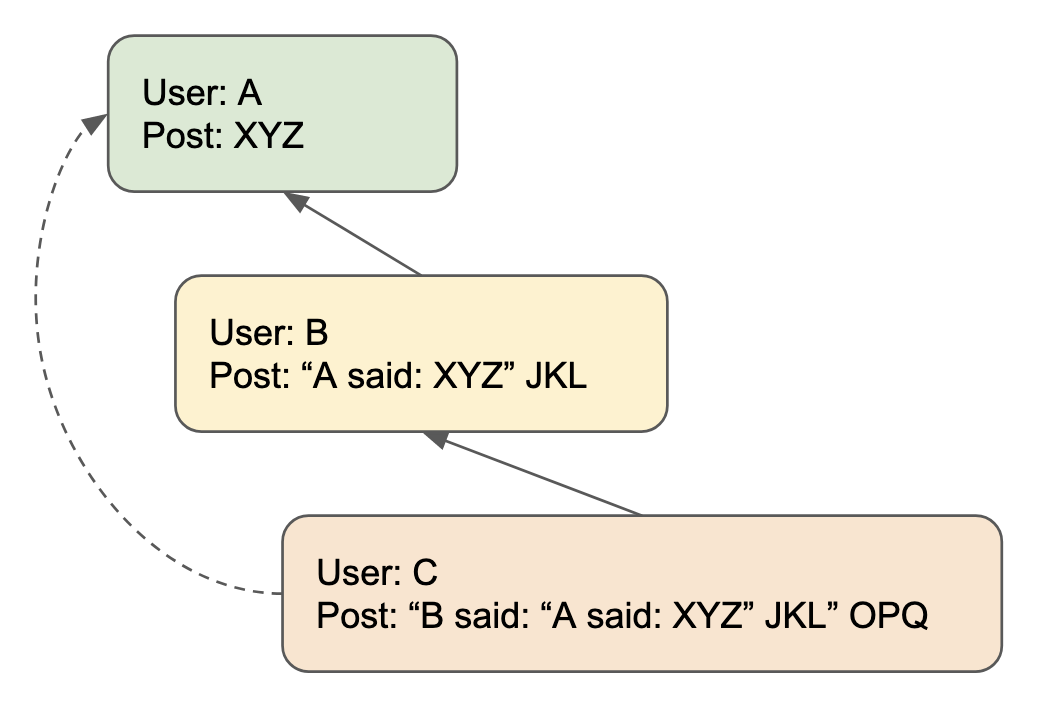}
    \caption{An interesting case in Hacker Forums dataset.}
    \label{fig:HFsample}
\end{figure}

In the Hacker Forums dataset, two hacker forums have a feature to quote the referencing post in the same thread. However, the quote feature catches not only the referencing post but also the referencing post's referencing post if it has. Fig.~\ref{fig:HFsample} shows the example of this case. User C replied (referenced) User B post, however, the User B referenced User A post as well. Then, both NPP and NPP-IP models predicted the pair of User C post and User A post are direct reply relationship since User C post contains User A post content through referencing User B post. We observed many this false positive cases, and this type of error effected some performance in the Hacker Forums dataset. Some pre-process to remove reference's reference post content will be needed to solve this issue.

\section{Conclusion}

Predicting thread structures within cybersecurity forums is a crucial component in  defining key social networks used to identify prominent users who provide useful information. Identifying these users can facilitate prediction and prevention of future cyber incidents and attacks.

A prompt-based learning model called Next Paragraph Prediction with Instructional Prompting (NPP-IP) for predicting thread structures across different cybersecruity topics was introduced. The method was evaluated using two different datasets and compared against several well known methods. The results show that the NPP-IP method had considerable improvement over existing methods, achieving the highest F1 score across different real world hacker forum datasets.

In the future work, we plan to train the language models with cybersecurity related data for adapting cybersecurity context to the models to improve the performance. In addition, we will apply these methods to the social network analysis to replace from the assumption methods' networks to our context considered thread structures for comparing the advantage of the improved quality of the networks.


\section*{Ethical Considerations}
In this research, we use one dataset from the other work with the agreement of using the dataset for this research only. We created the Hacker Forum dataset where the data from CYR3CON (Cyber Security Works), and they already anonymized the site names and usernames. We have an agreement with CYR3CON to use the data for this research only, and not sharing the data in public. The Hacker Forums dataset has randomly picked 20 threads that have average 15.4 posts per thread from three English hacker forums. The dataset was annotated by four cybersecurity experts (employees of CYR3CON) in a week as a part of their jobs.
Our goal is to construct thread structure from unstructured forums. Further precautions taken include not identifying individuals (including not publishing usernames), and presenting results objectively. In addition, we use well-known publicly released language models, BERT and RoBERTa, for our experiments.

\section*{Limitations}
In this study, we have presented a new approach of predicting thread structure in cybersecurity forums. However, there are several limitations. First, there is a challenge of accessibility of datasets, especially (public) sensitive. Many of the previous works used hacker forums as their datasets, however, most of them are not published due to copyright and other restrictions. Kashihara et al.~\cite{kashihara2020social} provided the Reddit dataset for this study only. In addition, CYR3CON provided us with raw hacker forums data for this research, and this data will not be published. Second, there is a language limitation. To compare our approach with the related work~\cite{kashihara2020social} on the same dataset, we need to stick with English since they used Reddit threads in English for their dataset. There are many hacker forums in not only English but also other languages, however, we selected 20 hacker forum threads from three English hacker forums for our evaluation since we fine-tuned English BERT and RoBERTa models for our evaluation. Lastly, another limitation is the size of ground truth. The original Reddit dataset by~\cite{kashihara2020social} used the Reddit tree structure that shows the reply relationship of each pair of posts that is the direct response as the ground truth. However, many hacker forums we used for evaluation do not have such a structure, and human experts needed to check every combination of post pairs under a thread to determine the relationship of each pair. Thus, we use the 20 threads from three English hacker forums annotated by human experts for the ground truth. 

\bibliography{ThesisRef}

\begin{thebibliography}{10}
\providecommand{\url}[1]{\texttt{#1}}
\providecommand{\urlprefix}{URL }

\bibitem{almukaynizi2017predicting}
Almukaynizi, M., Grimm, A., Nunes, E., Shakarian, J., Shakarian, P.: Predicting
  cyber threats through hacker social networks in darkweb and deepweb forums.
  In: Proceedings of the 2017 International Conference of The Computational
  Social Science Society of the Americas. p.~12. ACM (2017)

\bibitem{techtarget-article2021}
Culafi, A.: Ninety percent of dark web hacking forum posts come from buyers.
  \url{shorturl.at/fEGW4} (2021), accessed: 2022-04-01

\bibitem{BERT-DevlinCLT19}
Devlin, J., Chang, M., Lee, K., Toutanova, K.: {BERT:} pre-training of deep
  bidirectional transformers for language understanding. In: Proceedings of the
  2019 Conference of the North American Chapter of the Association for
  Computational Linguistics: Human Language Technologies, {NAACL-HLT} 2019,
  Minneapolis, MN, USA, June 2-7, 2019, Volume 1 (Long and Short Papers). pp.
  4171--4186 (2019), \url{https://aclweb.org/anthology/papers/N/N19/N19-1423/}

\bibitem{acl/ElsnerC08}
Elsner, M., Charniak, E.: You talking to me? {A} corpus and algorithm for
  conversation disentanglement. In: McKeown, K.R., Moore, J.D., Teufel, S.,
  Allan, J., Furui, S. (eds.) {ACL} 2008, Proceedings of the 46th Annual
  Meeting of the Association for Computational Linguistics, June 15-20, 2008,
  Columbus, Ohio, {USA}. pp. 834--842. The Association for Computer Linguistics
  (2008), \url{https://aclanthology.org/P08-1095/}

\bibitem{falcon2020framework}
Falcon, W., Cho, K.: A framework for contrastive self-supervised learning and
  designing a new approach. arXiv preprint arXiv:2009.00104  (2020)

\bibitem{CSS2021}
Fox, J.: Cybersecurity statistics 2021.
  \url{https://www.cobalt.io/blog/cybersecurity-statistics-2021} (2021),
  accessed: 2022-04-01

\bibitem{isi/FuAC07}
Fu, T., Abbasi, A., Chen, H.: Interaction coherence analysis for dark web
  forums. In: {IEEE} International Conference on Intelligence and Security
  Informatics, {ISI} 2007, New Brunswick, New Jersey, USA, May 23-24, 2007,
  Proceedings. pp. 342--349 (2007),
  \url{https://doi.org/10.1109/ISI.2007.379495}

\bibitem{goel2011cyberwarfare}
Goel, S.: Cyberwarfare: connecting the dots in cyber intelligence.
  Communications of the ACM  54(8),  132--140 (2011)

\bibitem{HalderKS19}
Halder, K., Kan, M., Sugiyama, K.: Predicting helpful posts in open-ended
  discussion forums: {A} neural architecture. In: Proceedings of the 2019
  Conference of the North American Chapter of the Association for Computational
  Linguistics: Human Language Technologies, {NAACL-HLT} 2019, Minneapolis, MN,
  USA, June 2-7, 2019, Volume 1 (Long and Short Papers). pp. 3148--3157 (2019),
  \url{https://www.aclweb.org/anthology/N19-1318/}

\bibitem{naacl/JiangCCW18}
Jiang, J., Chen, F., Chen, Y., Wang, W.: Learning to disentangle interleaved
  conversational threads with a siamese hierarchical network and similarity
  ranking. In: Walker, M.A., Ji, H., Stent, A. (eds.) Proceedings of the 2018
  Conference of the North American Chapter of the Association for Computational
  Linguistics: Human Language Technologies, {NAACL-HLT} 2018, New Orleans,
  Louisiana, USA, June 1-6, 2018, Volume 1 (Long Papers). pp. 1812--1822.
  Association for Computational Linguistics (2018)

\bibitem{isi/JohnsenF20}
Johnsen, J.W., Franke, K.: Identifying proficient cybercriminals through text
  and network analysis. In: {IEEE} International Conference on Intelligence and
  Security Informatics, {ISI} 2020, Arlington, VA, USA, November 9-10, 2020.
  pp. 1--7. {IEEE} (2020), \url{https://doi.org/10.1109/ISI49825.2020.9280523}

\bibitem{kashihara2020social}
Kashihara, K., Shakarian, J., Baral, C.: Social structure construction from the
  forums using interaction coherence. In: Proceedings of the Future
  Technologies Conference. pp. 830--843. Springer (2020)

\bibitem{sigkdd/LHuillierARA10}
L'Huillier, G., {\'{A}}lvarez, H., R{\'{\i}}os, S.A., Aguilera, F.: Topic-based
  social network analysis for virtual communities of interests in the dark web.
  {SIGKDD} Explorations  12(2),  66--73 (2010),
  \url{https://doi.org/10.1145/1964897.1964917}

\bibitem{MarinSS18}
Marin, E., Shakarian, J., Shakarian, P.: Mining key-hackers on darkweb forums.
  In: 1st International Conference on Data Intelligence and Security, {ICDIS}
  2018, South Padre Island, TX, USA, April 8-10, 2018. pp. 73--80 (2018),
  \url{https://doi.org/10.1109/ICDIS.2018.00018}

\bibitem{ijcnlp/MehriC17}
Mehri, S., Carenini, G.: Chat disentanglement: Identifying semantic reply
  relationships with random forests and recurrent neural networks. In: Kondrak,
  G., Watanabe, T. (eds.) Proceedings of the Eighth International Joint
  Conference on Natural Language Processing, {IJCNLP} 2017, Taipei, Taiwan,
  November 27 - December 1, 2017 - Volume 1: Long Papers. pp. 615--623. Asian
  Federation of Natural Language Processing (2017),
  \url{https://aclanthology.org/I17-1062/}

\bibitem{corr/abs-2109-07830}
Mishra, S., Khashabi, D., Baral, C., Choi, Y., Hajishirzi, H.: Reframing
  instructional prompts to gptk's language. CoRR  abs/2109.07830 (2021),
  \url{https://arxiv.org/abs/2109.07830}

\bibitem{corr/abs-2104-08773}
Mishra, S., Khashabi, D., Baral, C., Hajishirzi, H.: Natural instructions:
  Benchmarking generalization to new tasks from natural language instructions.
  CoRR  abs/2104.08773 (2021), \url{https://arxiv.org/abs/2104.08773}

\bibitem{cc-report2016}
Morgan, S.: Hackerpocalypse cybercrime report 2016.
  \url{https://cybersecurityventures.com/hackerpocalypse-cybercrime-report-2016/}
  (2016), accessed: 2022-04-01

\bibitem{paszke2017automatic}
Paszke, A., Gross, S., Chintala, S., Chanan, G., Yang, E., DeVito, Z., Lin, Z.,
  Desmaison, A., Antiga, L., Lerer, A.: Automatic differentiation in pytorch.
  In: NIPS-W (2017)

\bibitem{eurosp/PeteHCB20}
Pete, I., Hughes, J., Chua, Y.T., Bada, M.: A social network analysis and
  comparison of six dark web forums. In: {IEEE} European Symposium on Security
  and Privacy Workshops, EuroS{\&}P Workshops 2020, Genoa, Italy, September
  7-11, 2020. pp. 484--493. {IEEE} (2020),
  \url{https://doi.org/10.1109/EuroSPW51379.2020.00071}

\bibitem{phillips2015extracting}
Phillips, E., Nurse, J.R., Goldsmith, M., Creese, S.: Extracting social
  structure from darkweb forums  (2015)

\bibitem{chi/ReynoldsM21}
Reynolds, L., McDonell, K.: Prompt programming for large language models:
  Beyond the few-shot paradigm. In: Kitamura, Y., Quigley, A., Isbister, K.,
  Igarashi, T. (eds.) {CHI} '21: {CHI} Conference on Human Factors in Computing
  Systems, Virtual Event / Yokohama Japan, May 8-13, 2021, Extended Abstracts.
  pp. 314:1--314:7. {ACM} (2021), \url{https://doi.org/10.1145/3411763.3451760}

\bibitem{SarkarASS18}
Sarkar, S., Almukaynizi, M., Shakarian, J., Shakarian, P.: Predicting
  enterprise cyber incidents using social network analysis on the darkweb
  hacker forums. In: 2018 International Conference on Cyber Conflict, CyCon
  {U.S.} 2018, Washington, DC, USA, November 14-15, 2018. pp. 1--7 (2018),
  \url{https://cyberdefensereview.army.mil/Portals/6/Documents/CyConUS18\%20Conference\%20Papers/Session2-Paper2.pdf}

\bibitem{naacl/ScaoR21}
Scao, T.L., Rush, A.M.: How many data points is a prompt worth? In: Toutanova,
  K., Rumshisky, A., Zettlemoyer, L., Hakkani{-}T{\"{u}}r, D., Beltagy, I.,
  Bethard, S., Cotterell, R., Chakraborty, T., Zhou, Y. (eds.) Proceedings of
  the 2021 Conference of the North American Chapter of the Association for
  Computational Linguistics: Human Language Technologies, {NAACL-HLT} 2021,
  Online, June 6-11, 2021. pp. 2627--2636. Association for Computational
  Linguistics (2021), \url{https://doi.org/10.18653/v1/2021.naacl-main.208}

\bibitem{emnlp/SchickS21}
Schick, T., Sch{\"{u}}tze, H.: Few-shot text generation with natural language
  instructions. In: Moens, M., Huang, X., Specia, L., Yih, S.W. (eds.)
  Proceedings of the 2021 Conference on Empirical Methods in Natural Language
  Processing, {EMNLP} 2021, Virtual Event / Punta Cana, Dominican Republic,
  7-11 November, 2021. pp. 390--402. Association for Computational Linguistics
  (2021), \url{https://doi.org/10.18653/v1/2021.emnlp-main.32}

\bibitem{emnlp/TamMBSR21}
Tam, D., Menon, R.R., Bansal, M., Srivastava, S., Raffel, C.: Improving and
  simplifying pattern exploiting training. In: Moens, M., Huang, X., Specia,
  L., Yih, S.W. (eds.) Proceedings of the 2021 Conference on Empirical Methods
  in Natural Language Processing, {EMNLP} 2021, Virtual Event / Punta Cana,
  Dominican Republic, 7-11 November, 2021. pp. 4980--4991. Association for
  Computational Linguistics (2021),
  \url{https://doi.org/10.18653/v1/2021.emnlp-main.407}

\bibitem{ptsecurity-CHS2021}
Technologies, P.: Custom hacking services.
  \url{https://www.ptsecurity.com/ww-en/analytics/custom-hacking-services/}
  (2021), accessed: 2022-04-01

\bibitem{emnlp/WolfDSCDMCRLFDS20}
Wolf, T., Debut, L., Sanh, V., Chaumond, J., Delangue, C., Moi, A., Cistac, P.,
  Rault, T., Louf, R., Funtowicz, M., Davison, J., Shleifer, S., von Platen,
  P., Ma, C., Jernite, Y., Plu, J., Xu, C., Scao, T.L., Gugger, S., Drame, M.,
  Lhoest, Q., Rush, A.M.: Transformers: State-of-the-art natural language
  processing. In: Liu, Q., Schlangen, D. (eds.) Proceedings of the 2020
  Conference on Empirical Methods in Natural Language Processing: System
  Demonstrations, {EMNLP} 2020 - Demos, Online, November 16-20, 2020. pp.
  38--45. Association for Computational Linguistics (2020),
  \url{https://doi.org/10.18653/v1/2020.emnlp-demos.6}

\bibitem{emnlp/YuJ20}
Yu, T., Joty, S.R.: Online conversation disentanglement with pointer networks.
  In: Webber, B., Cohn, T., He, Y., Liu, Y. (eds.) Proceedings of the 2020
  Conference on Empirical Methods in Natural Language Processing, {EMNLP} 2020,
  Online, November 16-20, 2020. pp. 6321--6330. Association for Computational
  Linguistics (2020), \url{https://doi.org/10.18653/v1/2020.emnlp-main.512}

\bibitem{zhang2021identifying}
Zhang, P., Qi, Y., Li, Y., Ya, J., Wang, X., Liu, T., Shi, J.: Identifying
  reply relationships from telegram groups using multi-features fusion. In:
  2021 IEEE Sixth International Conference on Data Science in Cyberspace (DSC).
  pp. 321--327. IEEE (2021)

\bibitem{aaai/ZhuNWNX20}
Zhu, H., Nan, F., Wang, Z., Nallapati, R., Xiang, B.: Who did they respond to?
  conversation structure modeling using masked hierarchical transformer. In:
  The Thirty-Fourth {AAAI} Conference on Artificial Intelligence, {AAAI} 2020,
  The Thirty-Second Innovative Applications of Artificial Intelligence
  Conference, {IAAI} 2020, The Tenth {AAAI} Symposium on Educational Advances
  in Artificial Intelligence, {EAAI} 2020, New York, NY, USA, February 7-12,
  2020. pp. 9741--9748. {AAAI} Press (2020),
  \url{https://ojs.aaai.org/index.php/AAAI/article/view/6524}

\end{thebibliography}
\bibliographystyle{splncs03}
\end{document}